\documentclass[10pt,letterpaper]{article}

\usepackage{ccn}
\usepackage{pslatex}
\usepackage{apacite}
\usepackage{tikz} % some conferences restrict the usage of tikz for their submissions; please check in advance and go with a high-res screenshot instead
\usepackage{stfloats}
\usetikzlibrary{spy}
\usepackage{pgfplots}
\usetikzlibrary{spy,calc}
\usepackage{float}

\newif\ifblackandwhitecycle
\gdef\patternnumber{0}

\pgfkeys{/tikz/.cd,
    zoombox paths/.style={
        draw=orange,
        very thick
    },
    black and white/.is choice,
    black and white/.default=static,
    black and white/static/.style={ 
        draw=white,   
        zoombox paths/.append style={
            draw=white,
            postaction={
                draw=black,
                loosely dashed
            }
        }
    },
    black and white/static/.code={
        \gdef\patternnumber{1}
    },
    black and white/cycle/.code={
        \blackandwhitecycletrue
        \gdef\patternnumber{1}
    },
    black and white pattern/.is choice,
    black and white pattern/0/.style={},
    black and white pattern/1/.style={    
            draw=white,
            postaction={
                draw=black,
                dash pattern=on 2pt off 2pt
            }
    },
    black and white pattern/2/.style={    
            draw=white,
            postaction={
                draw=black,
                dash pattern=on 4pt off 4pt
            }
    },
    black and white pattern/3/.style={    
            draw=white,
            postaction={
                draw=black,
                dash pattern=on 4pt off 4pt on 1pt off 4pt
            }
    },
    black and white pattern/4/.style={    
            draw=white,
            postaction={
                draw=black,
                dash pattern=on 4pt off 2pt on 2 pt off 2pt on 2 pt off 2pt
            }
    },
    zoomboxarray inner gap/.initial=5pt,
    zoomboxarray columns/.initial=2,
    zoomboxarray rows/.initial=2,
    subfigurename/.initial={},
    figurename/.initial={zoombox},
    zoomboxarray/.style={
        execute at begin picture={
            \begin{scope}[
                spy using outlines={%
                    zoombox paths,
                    width=\imagewidth / \pgfkeysvalueof{/tikz/zoomboxarray columns} - (\pgfkeysvalueof{/tikz/zoomboxarray columns} - 1) / \pgfkeysvalueof{/tikz/zoomboxarray columns} * \pgfkeysvalueof{/tikz/zoomboxarray inner gap} -\pgflinewidth,
                    height=\imageheight / \pgfkeysvalueof{/tikz/zoomboxarray rows} - (\pgfkeysvalueof{/tikz/zoomboxarray rows} - 1) / \pgfkeysvalueof{/tikz/zoomboxarray rows} * \pgfkeysvalueof{/tikz/zoomboxarray inner gap}-\pgflinewidth,
                    magnification=3,
                    every spy on node/.style={zoombox paths},
                    every spy in node/.style={zoombox paths}
                }
            ] 
        },
        execute at end picture={\end{scope}},
        spymargin/.initial=0.5em,
        zoomboxes xshift/.initial=1,
        zoomboxes right/.code=\pgfkeys{/tikz/zoomboxes xshift=1},
        zoomboxes left/.code=\pgfkeys{/tikz/zoomboxes xshift=-1},
        zoomboxes yshift/.initial=0,
        zoomboxes above/.code={
            \pgfkeys{/tikz/zoomboxes yshift=1},
            \pgfkeys{/tikz/zoomboxes xshift=0}
        },
        zoomboxes below/.code={
            \pgfkeys{/tikz/zoomboxes yshift=-1},
            \pgfkeys{/tikz/zoomboxes xshift=0}
        },
        caption margin/.initial=4ex,
    },
    adjust caption spacing/.code={},
    image container/.style={
        inner sep=0pt,
        at=(image.north),
        anchor=north,
        adjust caption spacing
    },
    zoomboxes container/.style={
        inner sep=0pt,
        at=(image.north),
        anchor=north,
        name=zoomboxes container,
        xshift=\pgfkeysvalueof{/tikz/zoomboxes xshift}*(\imagewidth+\pgfkeysvalueof{/tikz/spymargin}),
        yshift=\pgfkeysvalueof{/tikz/zoomboxes yshift}*(\imageheight+\pgfkeysvalueof{/tikz/spymargin}+\pgfkeysvalueof{/tikz/caption margin}),
        adjust caption spacing
    },
    calculate dimensions/.code={
        \pgfpointdiff{\pgfpointanchor{image}{south west} }{\pgfpointanchor{image}{north east} }
        \pgfgetlastxy{\imagewidth}{\imageheight}
        \global\let\imagewidth=\imagewidth
        \global\let\imageheight=\imageheight
        \gdef\columncount{1}
        \gdef\rowcount{1}
        
    },
    image node/.style={
        inner sep=0pt,
        name=image,
        anchor=south west,
        append after command={
            [calculate dimensions]
            node [image container,subfigurename=\pgfkeysvalueof{/tikz/figurename}-image] {\phantomimage}
            node [zoomboxes container,subfigurename=\pgfkeysvalueof{/tikz/figurename}-zoom] {\phantomimage}
        }
    },
    color code/.style={zoombox paths/.append style={draw=#1}},
    connect zoomboxes/.style={spy connection path={\draw[draw=none,zoombox paths] (tikzspyonnode) -- (tikzspyinnode);}},
    help grid code/.code={
        \begin{scope}[
                x={(image.south east)},
                y={(image.north west)},
                font=\footnotesize,
                help lines,
                overlay
            ]
            \foreach \x in {0,1,...,9} { 
                \draw(\x/10,0) -- (\x/10,1);
                \node [anchor=north] at (\x/10,0) {0.\x};
            }
            \foreach \y in {0,1,...,9} {
                \draw(0,\y/10) -- (1,\y/10);                        \node [anchor=east] at (0,\y/10) {0.\y};
            }
        \end{scope}    
    },
    help grid/.style={append after command={[help grid code]}},
}

\newcommand\phantomimage{\phantom{\rule{\imagewidth}{\imageheight}}}

\newcommand\zoombox[2][]{
    \begin{scope}[zoombox paths]
        \pgfmathsetmacro\xpos{
            (\columncount-1)*(\imagewidth / \pgfkeysvalueof{/tikz/zoomboxarray columns} + \pgfkeysvalueof{/tikz/zoomboxarray inner gap} / \pgfkeysvalueof{/tikz/zoomboxarray columns} ) + \pgflinewidth
        }
        \pgfmathsetmacro\ypos{
            (\rowcount-1)*( \imageheight / \pgfkeysvalueof{/tikz/zoomboxarray rows} + \pgfkeysvalueof{/tikz/zoomboxarray inner gap} / \pgfkeysvalueof{/tikz/zoomboxarray rows} ) + 0.5*\pgflinewidth
        }
        \edef\dospy{\noexpand\spy [
            #1,
            zoombox paths/.append style={black and white pattern=\patternnumber},
            every spy on node/.append style={#1},
            x=\imagewidth,
            y=\imageheight
        ] on (#2) in node [anchor=north west] at ($(zoomboxes container.north west)+(\xpos pt,-\ypos pt)$);}
        \dospy
        \pgfmathtruncatemacro\pgfmathresult{ifthenelse(\columncount==\pgfkeysvalueof{/tikz/zoomboxarray columns},\rowcount+1,\rowcount)}
        \global\let\rowcount=\pgfmathresult
        \pgfmathtruncatemacro\pgfmathresult{ifthenelse(\columncount==\pgfkeysvalueof{/tikz/zoomboxarray columns},1,\columncount+1)}
        \global\let\columncount=\pgfmathresult
        \ifblackandwhitecycle
            \pgfmathtruncatemacro{\newpatternnumber}{\patternnumber+1}
            \global\edef\patternnumber{\newpatternnumber}
        \fi
    \end{scope}
}
\pgfplotsset{compat=1.18}

\renewcommand{\thefootnote}{\ifcase\value{footnote}\or*\else\arabic{footnote}\fi}

\title{EEG-Features for Generalized Deepfake Detection}
\author{{\large \bf Anonymous}}
\author{{\large \bf Arian Beckmann\thanks{Equal Contribution.}\textsuperscript{1} (arian.beckmann@hhi.fraunhofer.de)} \\
  \AND {\large \bf Tilman Stephani\footnotemark[1]\textsuperscript{2} (stephani@cbs.mpg.de)} \\
  \AND {\large \bf Felix Klotzsche\textsuperscript{2} (klotzsche@cbs.mpg.de)} \\
  \AND {\large \bf Yonghao Chen\textsuperscript{2} (cheny@cbs.mpg.de)} \\
  \AND {\large \bf Simon M.\ Hofmann\textsuperscript{2} (simon.hofmann@cbs.mpg.de)} \\
  \AND {\large \bf Arno Villringer\textsuperscript{2} (villringer@cbs.mpg.de)} \\
  \AND {\large \bf Michael Gaebler\textsuperscript{2} (gaebler@cbs.mpg.de)} \\
  \AND {\large \bf Vadim Nikulin\textsuperscript{2} (nikulin@cbs.mpg.de)} \\
  \AND {\large \bf Sebastian Bosse\textsuperscript{1} (sebastian.bosse@hhi.fraunhofer.de)} \\
  \AND {\large \bf Peter Eisert\textsuperscript{1,3} (peter.eisert@hhi.fraunhofer.de)} \\
  \AND {\large \bf Anna Hilsmann\textsuperscript{1} (anna.hilsmann@hhi.fraunhofer.de)}
  }

\begin{document}

\maketitle

{\small
\noindent
\textsuperscript{1}Fraunhofer Heinrich-Hertz-Institute, Einsteinufer 37, 10587 Berlin, Germany\\
\textsuperscript{2}Max Planck Institute for Human Cognitive and Brain Sciences, Stephanstraße 1A, 04103 Leipzig, Germany\\
\textsuperscript{3}Humboldt-Universität zu Berlin, Unter den Linden 6, 10099 Berlin, Germany \\
}

\section{Abstract}
{
\bf
Since the advent of Deepfakes in digital media, the development of robust and reliable detection mechanism is urgently called for. In this study, we explore a novel approach to Deepfake detection by utilizing electroencephalography (EEG) measured from the neural processing of a human participant who viewed and categorized Deepfake stimuli from the FaceForensics++ datset. These measurements serve as input features to a binary support vector classifier, trained to discriminate between real and manipulated facial images. We examine whether EEG data can inform Deepfake detection and also if it can provide a generalized representation capable of identifying Deepfakes beyond the training domain. Our preliminary results indicate that human neural processing signals can be successfully integrated into Deepfake detection frameworks and hint at the potential for a generalized neural representation of artifacts in computer generated faces. Moreover, our study provides next steps towards the understanding of how digital realism is embedded in the human cognitive system, possibly enabling the development of more realistic digital avatars in the future.

% The abstract should be identical to the text version submitted in the webform and should not exceed 1,500 characters, including spaces and any special characters. The abstract should thus be relatively short. Aim for 150 words.
% Max length is 200 words. Arbitrarily long German words like "Donaudampfschiffartskapit\"an" are not encouraged.
% CCN has an interdisciplinary audience. Hence a good abstract should
% (a) give context about what the problem is and why it matters 
% (b) give the contents and explain what was done and what was found
% (c) give a clear conclusion including what we learned and how it changes 
% the way we think about the universe.
% And because Konrad is writing this, he can not avoid shamelessly plugging
% his writing guide:
% \url{goo.gl/vC8tvf} See you at CCN.
}
\begin{quote}
\small
\textbf{Keywords:} 
EEG; Deepfake; perception; realism
\end{quote}

\section{Introduction}
The pope wearing puffer jackets, Tom Cruise doing magic tricks on TikTok, Donald Trump being arrested by the police. Visual "proof" of these situations went viral on social media -- yet they never happened. Deepfake technology can readily delude a viewer's beliefs about what a certain person says, does, and looks like. To maintain the delineation between truth and lie, it is hence of paramount importance for modern society to be able to identify and counteract such Deepfake technologies. A common approach for developing Deepfake detectors involves training convolutional neural networks (CNNs) to detect the spatio-temporal artifacts appearing in Deepfakes \cite{roessler2019faceforensicspp, altfreezing}. Although they perform well on benchmark datasets, these detectors often struggle to generalize to new manipulation domains that omit unfamiliar artifacts \cite{beckmannFooling}. Another possibility may be to add the human in the detection loop, and, more specifically, take advantage of the high-dimensional information contained in neurophysiological measurements of the perceptual and cognitive processing of Deepfake stimuli that leads or does not lead to the subjective percept of a fake human face. Previous work demonstrated that fake videos can be discriminated from genuine ones if the observer is familiar with at least one of the displayed persons \cite{tauscher2021}. Moreover, \cite{moschel2022} demonstrated that GAN generated images can be decoded by people's neural activity. In this proof-of-concept study, we test whether human electroencephalography (EEG) can inform the detection of deepfaked faces and whether it allows -- in contrast to naively trained CNNs -- to generalize across different Deepfake generation methods.

\begin{figure*}[!b]\centering
\begin{tikzpicture}[zoomboxarray]
    \node [image node] { \includegraphics[width=0.316\textwidth]{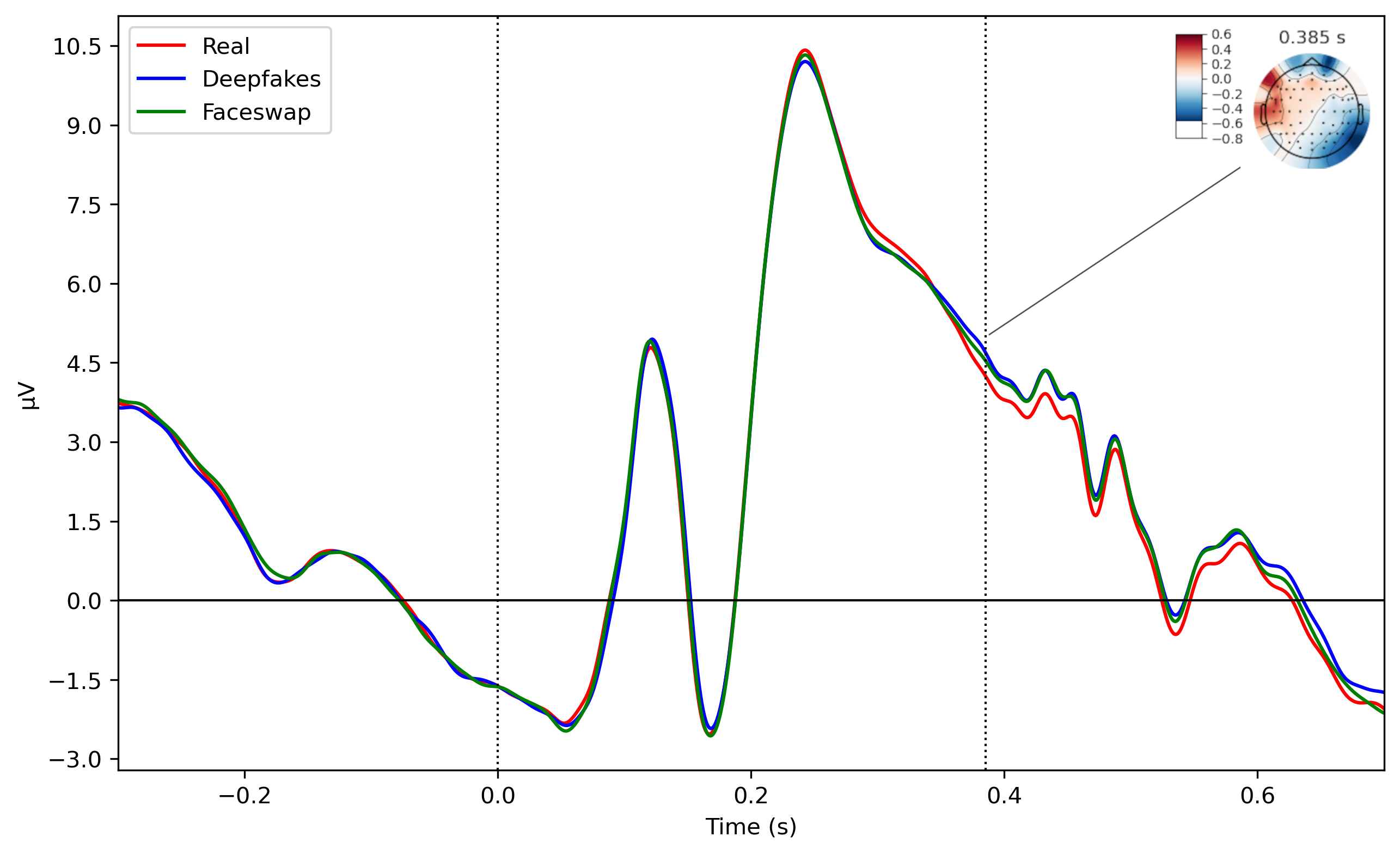} };
        \zoombox[magnification=5,color code=blue]{0.73,0.55}
        \zoombox[magnification=4,color code=yellow]{0.78,0.45}
        \zoombox[magnification=3,color code=red]{0.88,0.32}
        \zoombox[magnification=2.7,color code=green]{0.897,0.885}
\end{tikzpicture}
\caption{Mean EEG responses with respect to the three stimuli classes for electrode PO8 as well as the topography (across all electrodes) of the difference between fake and real images at 385 ms after stimulus onset (green box).}
\label{fig: erp}
\end{figure*}

\section{Methods}
\subsubsection{Stimuli}
For our stimulus set, we utilize the FaceForensics++ \cite{roessler2019faceforensicspp} benchmark dataset as it contains forged facial videos originating from various different manipulation methods. We use stimuli of two different fake methods, "Deepfakes" (DF)\footnote{The general term Deepfake originates from this seminal forgery method, to avoid confusion we refer to it solely as DF.} and "FaceSwap" (FS),  along with their respective original counterparts. We chose these two fake methods as they produce different characteristic artifacts that are not difficult to identify. Per category, 500 videos were selected of which we randomly selected 8 (16) frames as fake (real) images, in total resulting in 16,000 images with balanced fake/real labels. Note that we selected the videos such that, per category, 360 videos belong to the training set and 70 videos to the validation and testing sets respectively, as specified in \citeA{roessler2019faceforensicspp}.
\subsubsection{Experimental procedure} 
The images were presented in random order on a computer screen to a human observer (co-author, male, 22 years), while measuring EEG. Each image was centrally presented for 350 ms, followed by a blank screen with a fixation target for 350 ms. Subsequently, in a subset of the trials, the participant was tasked to indicate via button press whether he perceived the stimulus as real or fake (within 1000 ms). The tasks appear at random, in 12.5\% of the trials, as to avoid the participant expecting the task. In the remaining 87.5\% of trials, the experiment continued with stimulus presentation of the following trial. The whole experiment consisted of 160 blocks with 100 trials each, amounting to a total duration of around 4 hours measurement time.

\subsubsection{EEG setup and preprocessing}
EEG data were recorded from 63 Ag/AgCl electrodes at a sampling frequency of 1000 Hz with a NeurOne Tesla EEG system (Bittium, Oulu, Finland). A built-in band-pass filter between 0.16 and 250 Hz was used. Electrodes were placed according to the international 10-10 system, mounted in an elastic cap (EasyCap, Hersching, Germany). FCz served as reference and CPz as ground electrode. During offline processing, the EEG data were band-pass filtered between 0.5 and and 40 Hz, re-referenced to an average reference, and ICA served to remove eye blink, eye movement, and heart artifacts. Subsequently, the data were cut into epochs from -300 to +700 ms relative to stimulus onset, including a baseline correction from -200 to 0 ms. Epochs with values exceeding +-400 µV at any electrode were excluded from further analysis (n=4).

\subsubsection{Deepfake classification}
We construct the following experiment to analyze whether the recorded EEG data can inform Deepfake detection, particularly assessing the potential of a generalized artifact representation: For each video in each category, we average over all recorded trials to obtain a denoised sample. Thus, we are left with 1500 denoised samples, distributed evenly across the three categories -- "Deepfakes" (DF), "Faceswap" (FS) and "real". Note that for denoising real samples, we use 8 instead of 16 recorded trials to ensure a similar signal quality between real and fake samples. Then, we form training, validation and testing sets according to \citeA{roessler2019faceforensicspp}. We process the data in two different variations, resulting from an extensive ablation study. We denote these variants by \textbf{V1} and \textbf{V2} respectively. Both variations ignore the first 300 ms pre stimulus onset and, ultimately, merge the spatial and temporal dimensions before applying dimensionality reduction. For \textbf{V1}, we use all remaining data and reduce its dimensionality via PCA with 64 components. Concerning \textbf{V2}, we split the data with respect to the remaining 700 ms along the spatial and temporal dimensions into chunks of length 100 ms per electrode, resulting in 441 chunks (63 electrodes X 7 100 ms intervals). For each chunk, we train a separate binary support vector classifier (SVC) to discriminate between neural signals representing real or deepfaked stimuli. Subsequently, we evaluate the classifiers on the validation sets of the respective chunks. The top 100 chunks by validation F1-score were selected, consolidated and further reduced by ICA with 128 components.

The following training and evaluation process is performed separately for data pre-processed according to both \textbf{V1} and \textbf{V2}: We train an SVC (with default parameters in scikit-learn) to discern real and fakes on training data containing DF and "real" and evaluate it on the respective test set. Moreover, to test out-of-domain detection performance, we evaluate the classifier on the testing subset of FS. Likewise, we perform the experiment including FS instead of DF in the training set, while still testing on both fake subsets. The results of our experiments are shown in Table \ref{table:scores}. DF$\rightarrow$FS refers to the case in which the train set contains DF and "real" and the model is evaluated on the testing sets corresponding to FS and "real". The other columns follow the same logic.
\begin{table}[!t]
\begin{center} 
\caption{Macro F1-Scores for both variations on multiple train-test splits. Bold numbers highlight out-of-domain testing.} 
\label{table:scores} 
\vskip 0.12in
\begin{tabular}{c|cccc} 
\hline
Variation & DF$\rightarrow$ DF & DF$\rightarrow$ FS & FS$\rightarrow$ DF & FS$\rightarrow$ FS \\
\hline
\textbf{V1} & 0.62 & \textbf{0.58} & \textbf{0.59} & 0.61 \\
\textbf{V2} & 0.61 & \textbf{0.58} & \textbf{0.61} & 0.56 \\

\hline
\end{tabular} 
\end{center} 
\end{table}

\section{Results and Discussion}
The left-hand side of Figure \ref{fig: erp} shows the EEG responses averaged over the respective classes for electrode PO8. The magnified regions on the right-hand side show a significant difference between the responses to the real images and their manipulated counterparts (confirmed by cluster-based permutation testing). Additionally, the green box displays the topography of the difference in the responses to faked (DF and FS) and real images at 385 ms after stimulus onset. These descriptive results tentatively demonstrate that neural processing may contain a generalized representation of artificiality with respect to computer-generated faces. This interpretation obtains further support from the decoding results depicted in Table \ref{table:scores}. As can be seen in the third and fourth columns, the classifier is able to produce above chance level performance when confronted with fakes not seen during training. We check the significance of these results by permutation testing against chance-level with 10,000 repetitions for $p=0.05$. The resulting $p$-values are $.0309$ and $.0128$ for \textbf{V1}, as well as $.0277$ and $.0036$ for \textbf{V2} (same order as shown in Table \ref{table:scores}). Nonetheless, to further support our hypothesis, we aim to perform more experiments with a wider variety of Deepfakes and more participants in our future work.

\section{Conclusion}
In this pilot experiment, we not only demonstrated that features derived from EEG recordings can be used to detect Deepfakes, but also that these features can be utilized for out-of-domain fake detection -- hinting at the potential for a generalized representation of artifacts or uncanny content within neural processing signals. For subsequent experiments, we plan to include more high-quality images and to manually add a variety of artifacts, to enable more control and a broader analysis.

\section{Acknowledgments}
This work has received funding through the Max Planck-Fraunhofer collaboration project NeuroHum.

\bibliographystyle{apacite}

\setlength{\bibleftmargin}{.125in}
\setlength{\bibindent}{-\bibleftmargin}

\bibliography{ccn_style}

\end{document}